\documentclass[10pt,twocolumn]{article}

\usepackage[letterpaper,top=0.9in,bottom=0.9in,left=0.875in,right=0.875in,columnsep=0.25in,headheight=12pt,headsep=0.2in]{geometry}
\usepackage[utf8]{inputenc}
\usepackage[T1]{fontenc}
\usepackage{newtxtext}
\usepackage{newtxmath}
\usepackage[round,authoryear]{natbib}
\usepackage{hyperref}
\usepackage{url}
\usepackage{booktabs}
\usepackage{amsmath}
\usepackage{graphicx}
\usepackage{microtype}
\usepackage{xcolor}
\usepackage{tabularx}
\usepackage{xspace}
\usepackage{enumitem}
\usepackage{titlesec}
\usepackage{fancyhdr}
\usepackage{balance}
\usepackage{placeins}
\usepackage[font=small,labelfont=bf]{caption}

\definecolor{mtvagreen}{HTML}{129C80}
\definecolor{linkblue}{HTML}{174A8B}
\hypersetup{colorlinks=true,citecolor=linkblue,linkcolor=linkblue,urlcolor=linkblue}
\setlist{itemsep=2pt,topsep=0pt,parsep=0pt,leftmargin=1.5em}
\titleformat{\section}{\large\bfseries}{\thesection.}{0.45em}{}
\titleformat{\subsection}{\normalsize\bfseries}{\thesubsection.}{0.45em}{}
\titleformat{\paragraph}[runin]{\normalsize\bfseries}{}{0pt}{}
\titlespacing*{\section}{0pt}{15pt plus 4pt minus 3pt}{7pt plus 1pt}
\titlespacing*{\subsection}{0pt}{10.5pt plus 3pt minus 2pt}{4.5pt plus 1pt}
\titlespacing*{\paragraph}{0pt}{7pt plus 2pt minus 1pt}{0.6em}
\fancypagestyle{firstpage}{
  \fancyhf{}
  \fancyfoot[C]{\thepage}
  
}

\newcommand{\bench}{\textsc{MTVA-Bench}\xspace}

\newcommand{\model}[1]{\textsf{#1}}
\newcolumntype{Y}{>{\raggedright\arraybackslash}X}
\newcolumntype{P}[1]{>{\raggedright\arraybackslash}p{#1}}

\begin{document}
\twocolumn[
\begin{@twocolumnfalse}
  \vspace*{-1.25em}
  \noindent\rule{\textwidth}{0.7pt}
  \vspace{0.9em}
  \begin{center}
    {\LARGE\bfseries MTVA-Bench: Evaluating the Language Model\\[0.18em]Inside Cascaded Voice Agents\par}
  \end{center}
  \vspace{0.55em}
  \noindent\rule{\textwidth}{0.7pt}
  \vspace{0.75em}
  \begin{center}
    {\normalsize Pritish Mishra, Ishaan Kumar, Akshat Mandoli, Sudarshan Kamath\par}
    \vspace{0.45em}
    {\normalsize\bfseries Smallest AI\par}
    \vspace{0.15em}
  \end{center}
  \vspace{1.0em}
\end{@twocolumnfalse}
]
\thispagestyle{firstpage}

\begin{abstract}
  Generally, most voice agents are cascaded systems, i.e., an ASR model transcribes the caller's audio, a language model reads the transcript and decides what to say and which backend tools to call, and a TTS model speaks the reply. Nearly all of the decision making happens in the language model, but existing evaluations measure it either too broadly or too narrowly. End-to-end voice benchmarks score the full pipeline, so recognition errors and model errors mix into a single number. LLM benchmarks isolate the model but they do not evaluate what makes real phone calls hard, such as transcription issues, caller's voice being split across messages and the requirement that replies follow the language and script specified. We introduce the \textbf{Multi-Turn Voice Agent Benchmark (MTVA-Bench)}, which evaluates the language model on the same conditions it faces inside a cascaded system. The caller is played by an LLM following a set of rubrics and tool calls are answered by a mock backend which responds to the arguments the model actually sent. The benchmark contains 49 agents working across 490 reviewed scenarios and supports 7 languages. Scoring is a combination of deterministic checks on tool calls with two LLM judges, one that scores scenario specific rules and one that grades conversation quality without access to the task. Both judges must cite specific messages from the transcript. Task and conversation scores are weighted equally, since a call can complete its task and still go badly for the caller. In a seven-model study, six of the models select the correct tool within 6.4 points of one another, but their overall scores span 24.4 points. Most of the gap comes from argument values, action ordering, rule compliance, and what the model says around its tool calls.
\end{abstract}

\section{Introduction}

A caller phones a courier company to move a delivery date, and the recognizer garbles one digit of the tracking number. The model behind the agent picks the right API, \texttt{reschedule\_delivery}, so a function-calling benchmark would count this call as a success. It also passes the garbled number, fires the call before the caller has agreed to the new date, gets an error back from the backend, and then tells the caller everything is confirmed. Four separate things went wrong on this call, and none of them show up in a metric that stops at the function name.

This paper is about measuring those failures. A production voice agent is usually a cascade. ASR turns the caller's audio into text, a language model reads the text and decides what to do, and TTS speaks the reply. Almost all of the real work falls on the model in the middle. It has to work out what the caller meant, remember what has already been said, judge whether identity was verified and consent was given, pick the backend action that is allowed at that moment, understand what a tool result means, and then say something useful about it. The caller never sees the backend, so their entire picture of what the company did comes from this one model.

A phone call also brings problems that a chat window does not have. Recognition is imperfect, so the text the model reads is not always what the caller said. Voice activity detection (VAD) decides when the caller has finished speaking, and it sometimes cuts one sentence into two or three messages. And TTS reads the model's output exactly as written, so a reply in the wrong script, or one with markdown in it, gets spoken to the caller as is. A model can be very good at text-chat tool use and bad at all three of these.

Existing benchmarks look at either too much of the pipeline or too little of it. End-to-end voice benchmarks score the whole system, and since recognition, the model, tool execution, and TTS all shape the same call, the score cannot say which one failed \citep{bogavelli2026evabench,ray2026tauvoice,meyer2026vamos}. Text-agent benchmarks measure the model on its own but drop the call, which removes exactly the work described above. Dialogue datasets measure state tracking and response generation \citep{budzianowski2018multiwoz,rastogi2020sgd}, stateful agent benchmarks add simulated users, tools, and policies \citep{yao2024taubench,barres2025tau2,lu2025toolsandbox}, and recent speech benchmarks evaluate spoken assistants directly \citep{chen2026voicebench,jain2025voiceagentbench}. We want something in between, a way to see how the language model behaves inside a voice loop without blaming it for errors made by audio components it does not control.

\bench does this by holding the model's production interface fixed and simulating everything on the other side of it. The model under test receives the agent's operating instructions, its tool schemas, and the caller's turns as post-ASR text. It replies with speech text and tool calls, exactly as it would in production. The caller is another LLM that improvises from a private brief. The brief fixes who the caller is, what they want, and which facts they hold, but the wording reacts to whatever the model says, and the harness rejects caller turns that jump ahead in the plan or reveal hidden information. A channel layer garbles or splits the caller's words in ways each scenario declares, which reproduces in text what ASR and VAD do to input. A mock backend holds a response queue for every tool and answers each call based on the arguments the model actually sent. A wrong argument produces a failure response instead of the happy path. Nothing about the audio itself is simulated or scored. Pronunciation, background noise, and latency never appear at the model's interface, so the benchmark never charges the model for them.

Three evaluators grade the saved call, and none of them sees everything. Program code checks the tool calls exactly. A scenario judge reads the call against rules written for that specific scenario. A conversation judge rates how the call went without ever learning what the task was. The two LLM judges must cite the index of a real message or tool event for every judgment they make, and any rule they claim was broken must be quoted verbatim from the agent's instructions.

This paper makes four contributions.

\begin{enumerate}
  \item A precise evaluation target. \bench tests the language model between ASR and TTS, keeping everything that survives that boundary in text form and deliberately excluding audio quality and timing.
  \item A controlled environment of 490 reviewed multi-turn scenarios over 49 full-length agent configurations in seven languages, in which the caller's knowledge, the hidden backend state, the tool behavior, and the grading rules are authored as separate objects.
  \item An evaluator split across three graders with separate views of the same saved call, where every LLM judgment must cite evidence that code verifies to exist.
  \item A seven-model study with paired uncertainty estimates, plus nine controlled probes of the evaluator itself. Six of the seven models pick the correct operation within 6.4 points of one another while their overall scores span 24.4 points.
\end{enumerate}

\begin{figure*}[t]
  \centering
  \includegraphics[width=\textwidth]{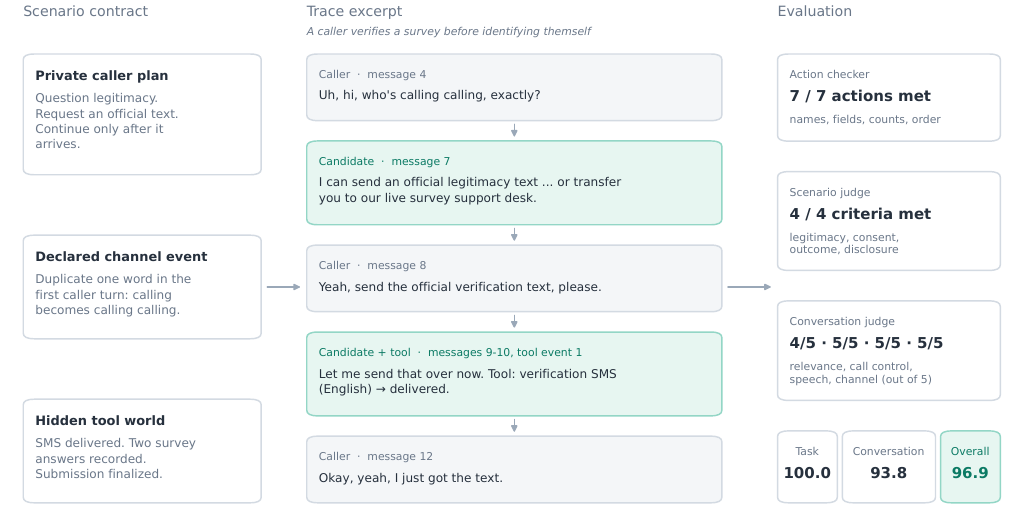}
  \caption{One scored \bench episode. The center panel shows an abridged excerpt from the saved \model{electron} trace for scenario \texttt{gov\_cs\_006\_agent\_0\_002}. The caller asks for verification before confirming identity, and the candidate sends a text only after obtaining consent. Counts and scores come from the full indexed trace.}
  \label{fig:sample-trace}
\end{figure*}

\section{What the benchmark measures}
\label{sec:design}

\subsection{The interface under test}

One episode is one complete simulated phone call, saved as an indexed trace and scored as a whole. Formally, an agent configuration $a$ has operating instructions $p_a$, a typed tool interface $\mathcal{T}_a$, and typed conditions that define when a call-ending tool has succeeded. A scenario $s$ has private caller knowledge $U_s$, hidden backend state $W_s$, a channel program $h_s$, executable action checks $A_s$, and judged criteria $C_s$. Running candidate $m$ produces a trace $\tau$, and the benchmark computes
\begin{equation}
  E(m,s,a) = \tfrac{1}{2}S_{\mathrm{task}}(\tau,A_s,C_s)
  + \tfrac{1}{2}S_{\mathrm{conversation}}(\tau,p_a,h_s).
  \label{eq:score-overview}
\end{equation}

The two halves ask different questions. The task half asks whether the right things happened, so it looks at the tools called, the arguments they carried, the state the call reached, and the rules it respected along the way. The conversation half asks what the call was like to sit through, so it looks at memory across turns, speech around tool calls, and whether the replies could be spoken as written. We keep the halves separate and weight them equally because each can hide the other. A polished apology can distract from a failed booking, and a completed booking does not mean the caller could follow the call.

We score whole episodes rather than single turns. A model that sends a wrong argument in turn three and speaks beautifully afterward has still failed the caller, and turn-level scores let that failure wash out. We also keep the caller's knowledge and the backend's state strictly apart. The caller knows her own account number but not what the database says about her account, so the model must learn the backend's truth by calling tools, the way it would in production.

The obligations of the voice channel survive in text form. Garbled words, split messages, required languages and scripts, spoken numbers, speech around tool calls, and an audible goodbye are all present in the trace and all scorable. Figure~\ref{fig:sample-trace} shows one scored episode alongside the scenario that produced it and the three grader views that consumed it.

\subsection{How a call runs}

The candidate model speaks first, since a phone agent greets the caller or introduces itself on an outbound call. Then the loop begins. The caller advances one intention from its brief and speaks. The channel renders those words, possibly garbled or split into fragments. The candidate answers with speech, tool calls, or both, and the backend answers each tool call from its queue. At caller turn $t$,
\begin{align}
  (u_t,D_t,q_t) &\leftarrow \mathrm{Caller}(U_s,H_t), \\
  (x_{t,1},\ldots,x_{t,k}) &\leftarrow \mathrm{Channel}_{h_s}(u_t), \\
  (z_t,c_t) &\leftarrow m(p_a,H_t,x_{t,1:k},\mathcal{T}_a), \\
  r_t &\leftarrow \mathrm{ToolWorld}(W_s,c_t,j_t),
\end{align}
where $u_t$ is the caller's intended utterance, $D_t$ the plan steps it newly expressed, $q_t$ its completion flag, $H_t$ the public conversation, $z_t$ and $c_t$ the candidate's speech and tool call, and $j_t$ counts prior calls to the selected tool. A candidate turn may chain up to ten tool calls before the caller speaks again. The episode ends on a successful call-ending tool result, on the caller deciding the call is over, or at thirty caller turns. Section~\ref{sec:simulation} explains why a failed call-ending tool does not end the episode.

Two small accommodations keep the harness from penalizing chat-template differences. Models whose templates require a user message before the first assistant turn get an empty one, and models that cannot emit speech and a tool call in the same message get a \texttt{concluding\_message} field on call-ending tools so they can still say goodbye.

\subsection{Design rules}

\paragraph{Failure attribution.}
A garbled word belongs to the channel program. A wrong tool argument belongs to the candidate. A failure response follows from the scenario's authored fixtures. The trace stores each of these separately, so when the model claims success after a failed call, the returned object is right there to check the claim against.

\paragraph{No reference answers.}
A service call has many valid phrasings and often more than one valid path, so we never compare against a reference reply. What is fixed is the state the call must reach and the rules it must respect on the way. The caller's plan fixes what enters the conversation, the tool fixtures fix what each action returns, and the action checks and criteria fix which paths count as success.

\paragraph{Voice obligations without audio.}
The candidate must repair recognition damage, join split messages, retain spoken values, speak around tool calls, and produce text in the required language and script. We do not grade pronunciation, tone of voice, background noise, or latency, because none of them reach the model's interface.

\paragraph{Not applicable rather than failed.}
When a trace never gives the model a chance to show a behavior, that criterion is marked not applicable and drops out of the score. A model should not earn channel-repair credit on a clean trace, and it should not lose points for a repair the episode never demanded.

An \bench score describes one candidate under one fixed caller, channel, backend, judge, and scoring version. It is not a universal ranking and it is not a safety certification.

\FloatBarrier
\section{The scenarios}
\label{sec:construction}

\subsection{What one scenario contains}

An example makes the pieces concrete. In one Gujarati scenario, a dairy cooperative's outbound agent calls a farmer to collect her milk estimate for the evening pickup. The caller simulator receives a private brief. The farmer will supply no milk because one cow has mastitis and the veterinary worker said to discard today's milk. She wants a short call, gives the reason only when asked, and if pressed to supply anyway she declines once and moves to end the call. The candidate model sees none of this. It sees the agent's 38,812-character operating document and 18 tool schemas. The mock backend is authored so that \texttt{log\_zero\_supply\_reason} returns a confirmation only when the model sends the right collection center, the evening shift, and the category \texttt{cattle\_health}, and returns \texttt{not\_recorded} for anything else. The graders hold the expected actions and the scenario's goal, which requires the zero estimate and its reason to be recorded without pressuring the farmer.

\begin{table}[t]
  \caption{Who sees which part of a scenario.}
  \label{tab:contract}
  \small
  \begin{tabularx}{\columnwidth}{@{}P{0.23\columnwidth}P{0.24\columnwidth}Y@{}}
    \toprule
    Scenario component & Visible to & Purpose \\
    \midrule
    Caller context, known and unknown facts, ordered plan steps & Caller simulator & Fixes what the caller wants, knows, and may reveal, without scripting lines \\
    Agent instructions, tools, public conversation & Candidate & Defines the operating policy and available actions \\
    Hidden backend state, ordered tool fixtures, defaults & Tool environment & Determines tool responses without exposing backend truth to caller or candidate \\
    Expected actions, criteria and anchors, prompt excerpts & Evaluators only & Defines success, partial progress, and rule compliance \\
    ASR events and VAD splits & Channel renderer and conversation judge & Creates declared input damage and records intended and observed text \\
    \bottomrule
  \end{tabularx}
\end{table}

Table~\ref{tab:contract} lists who sees what, and the separation matters in both directions. A caller that could read the backend state would quietly feed the model answers it should have had to earn through tools. A judge that could read the expected outcome would grade against what should have happened instead of what the trace shows.

Three typed objects carry the grading rules. A plan step has a trigger and a meaning, not a line to recite. An expected action names a tool and pins its arguments, call counts, and ordering with typed matchers. A criterion has an importance weight, a statement, and three written anchors describing what met, partial, and not met look like for that scenario. Criteria that depend on the agent's instructions carry exact excerpts of those instructions, and the loader verifies each excerpt against the hash of the full rendered prompt. A criterion cannot hold the agent to a rule it was never given.

\subsection{The corpus}

The bundle has 50 agent definitions, of which 49 are active with ten scenarios each. They cover finance, healthcare, public services, logistics, retail, travel, agriculture, education, and emergency intake. These are long operating documents rather than toy prompts. The median rendered instruction set is 30,834 characters, the range is 21,658 to 43,969, and the active agents declare 371 tools between them, from one to eighteen each. Table~\ref{tab:corpus} summarizes the composition.

\begin{table}[t]
  \caption{Composition of the evaluated bundle. Channel counts describe authored scenario conditions.}
  \label{tab:corpus}
  \footnotesize
  \begin{tabularx}{\columnwidth}{@{}P{0.27\columnwidth}rY@{}}
    \toprule
    Item & Count & Distribution \\
    \midrule
    Cases / active agents & 490 / 49 & 10 scenarios per agent \\
    Difficulty & 490 & 245 medium, 245 hard \\
    Languages & 7 & English 80, Hindi 60, five languages 70 each \\
    Caller plan steps & 4,065 & 4--12 per scenario \\
    Action checks & 2,249 & 1,866 required, 377 prohibited, 6 optional \\
    Task criteria & 1,621 & 2--6 per scenario \\
    Prompt excerpts & 1,245 & 0--8 per scenario \\
    Channel conditions & 53 & 10 light ASR, 43 natural VAD splits \\
    \bottomrule
  \end{tabularx}
\end{table}

\model{gpt-5.6-sol} generated the scenarios from the full agent prompts at high reasoning effort, and the authors reviewed every case. Mechanical audits check schemas, tool references, matcher types, action dependencies, caller isolation, and excerpt hashes, and a generation manifest records the parameters, seeds, and hashes behind the bundle. Review of this kind documents where the data came from. It does not provide independent labels or agreement between annotators.

Each case is authored in its language from the start rather than translated afterward. The bundle is balanced by difficulty but not by dialect, region, accessibility need, or real deployment traffic.

\section{The simulated call}
\label{sec:simulation}

\subsection{The caller}

The caller simulator receives its context, its reason for calling, the facts it knows, the facts it explicitly does not know, its ordered plan steps, its language, and the date and time. It never receives the backend state, the expected actions, or anything from the graders.

The plan is a list of steps, and each step is a trigger plus a meaning. In the dairy scenario one step reads, in effect, when the agent asks for the estimate, say that no milk will come, so the estimate is zero. We use meanings instead of scripted lines because the candidate's path is not fixed. It may ask valid questions in a different order, retry a failed action, or offer a permitted alternative, and a script rewards whichever path its author happened to anticipate while breaking on every other. The caller advances at most one step per turn, in order, and only when its trigger has come true.

The rest of the caller's behavior copies how people talk on the phone. A routine turn is one short sentence, usually 2 to 15 words. The caller answers one question at a time even when the agent asks three. It may need a moment to find an account number, and saying it will check can be its whole turn. It may invent harmless texture, a kettle or traffic, but never a business fact. If the agent asks something outside its known facts it says it does not know. It speaks its language in native script, and it writes what a recognizer would capture rather than how a document prints, so no markdown, no bullet lists, and long numbers grouped by breath.

Each caller turn is returned as typed JSON holding the spoken text, any newly expressed step, and a completion flag. The harness rejects empty turns, steps advanced out of order or twice, and any turn that leaks an internal field name into speech. A rejected turn gets one more attempt, a conversation that still fails is rerun up to three times, and nothing is ever saved half-finished.

Fixing the caller's facts and plan does not make the caller neutral. Its language ability and its habits from training still leak into every trace. Fair comparisons therefore run every candidate against the same caller model, prompt, and settings.

\subsection{The channel}

The channel layer decides what text the candidate observes. An authored ASR event drops, doubles, or substitutes a specific span at a specific turn. A lighter seeded mode changes one eligible word of the first caller turn, drawn from words of four letters or more and never from the caller's known values, which keeps the damage away from the exact fact the call depends on unless a scenario puts it there. Every event records what it changed and whether it applied.

VAD scenarios split a caller turn into at most three consecutive messages, and the split lands where a real pause would, inside a long identifier or between two halves of a thought. These cases test whether the model waits and reads the fragments as a single meaning. The candidate sees only the observed fragments. The intended text stays in the trace so a grader can later tell damage from the caller's own wording, but it is never shown to the model.

\subsection{The backend}

Every tool holds an ordered queue of authored responses. The $j$th call to a tool is matched against the $j$th entry, whose expected arguments are typed matchers over equality, presence, membership, numeric range, regular expressions, list containment, and object subsets. Matching arguments get the authored response. A mismatch or an unparseable call gets that entry's mismatch response, and a call beyond the queue gets the tool's default. Fixture values can reference the caller's facts or the hidden state, which is how the dairy scenario knows to accept \texttt{cattle\_health} and reject anything else. The backend therefore reacts to what the model actually sent while staying fully replayable.

Calling a call-ending tool does not by itself end the episode. The episode ends only when the returned object satisfies the agent's typed success condition, for example \texttt{end\_call} returning a status of \texttt{ended}. A failed booking, transfer, or record hands control back to the model, which must explain, retry where allowed, or fall back. This rule exists because ending the episode on the attempt would convert failures into completions. It also leaves hard evidence in the trace against whatever the model later claims about the outcome.

\subsection{The trace}

The saved trace is the only object anything is graded on. Every message keeps its index, role, content, and tool-call structure. Every caller turn keeps its intended text, observed fragments, expressed plan steps, and applied channel events. Every tool event keeps its parsed arguments, parse errors, matcher results, the returned object, and whether it ended the call. When a judge cites evidence, it cites these indices.

Validation runs before any scoring. References must resolve, matcher types must agree with the tool schemas, excerpts must match the prompt hash, and call-ending tools must declare success conditions. Component seeds derive from the run seed, so channel and backend behavior replays exactly. Provider calls replay only as far as the endpoint honors the recorded settings, which in practice varies.

\section{Scoring}
\label{sec:evaluation}

\begin{table*}[t]
  \caption{The ten diagnostic labels used in reports. Each names an observable question rather than a broad quality trait.}
  \label{tab:diagnostics}
  \centering
  \small
  \begin{tabularx}{0.97\textwidth}{@{}P{0.15\textwidth}P{0.17\textwidth}Y@{}}
    \toprule
    Pillar & Report label & Question represented \\
    \midrule
    Task & right operation & Was each required tool called, and was every prohibited one avoided? \\
    Task & right fields & Did the call carry the required argument values and no forbidden extras? \\
    Task & right order & Did calls happen in the required order and the required number of times? \\
    Task & requested state & Did the tool returns and the dialogue establish the state the caller asked for? \\
    Task & workflow progress & When completion was impossible, did the model reach the best supported next step? \\
    Task & operating rules & Were the consent, verification, disclosure, and retry rules of this case obeyed? \\
    Conversation & channel repair & Did the model recover a declared garbled word or join declared fragments? \\
    Conversation & live-call control & Did it retain answered facts, bridge tool calls audibly, and close the call properly? \\
    Conversation & speakable language & Could the emitted text go straight to TTS in the required language and script? \\
    Conversation & natural relevance & Did each spoken turn answer the moment directly, without padding or repetition? \\
    \bottomrule
  \end{tabularx}
\end{table*}

Exact tool behavior, task meaning, and the conduct of the call are three different claims, and we grade each with a different evaluator on a different slice of the evidence. This guards against anchoring. When everything sits in one judging prompt, a fluent transcript starts to sway the tool verdict, and knowledge of the intended outcome colors the rating of how the call felt. Each evaluator therefore gets the smallest view its decision needs.

\paragraph{The action checker.}
This evaluator is code, with no model involved. For every expected action it asserts presence, prohibited status, argument matches, forbidden extras, call-count bounds, and ordering against the typed tool events. When every assertion passes the action is labeled \texttt{met}. A call that exists in the right position but fails an argument or count check earns \texttt{partial}, and a missing, prohibited, or misordered call earns \texttt{not\_met}. Optional actions that were never used drop out entirely.

\paragraph{The scenario judge.}
An LLM handles what code cannot read off the arguments. It decides whether verification happened before disclosure, whether consent was given or merely assumed, whether the model's spoken claims match what the tools returned, and whether the model reached the best available next step when the request could not be fulfilled. The judge receives the agent's full instructions, the indexed transcript, the caller's known facts, the tool events, and the checker's raw facts with their verdict labels and importance weights stripped out, so it works from what happened rather than echoing verdicts or guessing which criteria matter most. It cannot revise the executable record.

Its output is validated by code. Every criterion must appear exactly once and in order, every applicable judgment must cite a message or tool event index that exists, and any claimed rule violation must quote the agent's instructions verbatim. A response that breaks the contract gets one retry and is then discarded. This check has a narrow reach. It proves the cited evidence exists and the output is well formed, but whether the judge read the evidence correctly is a separate question, and Section~\ref{sec:validation} takes that question up.

The judge may also flag up to five rule violations that no authored criterion covers. These stay unscored, because letting a judge invent new penalties after reading a trace would make scores incomparable across runs, but they are stored, and their volume travels with the results.

\paragraph{The conversation judge.}
A second LLM works from the transcript alone. It rates whether each reply addressed the moment, whether the model retained what it was told and kept the caller oriented through tool calls, whether its text was ready for TTS in the required language and script, and whether it recovered the input the channel garbled or split. This judge never sees the goal, the hidden state, the expected actions, or any task score. It reads the agent's instructions only to learn the required scripts, disclaimers, and language policy, so that mandated behavior is not mistaken for a defect.

Each dimension has five written anchors, and ownership rules assign each defect to exactly one dimension so one mistake cannot sink two ratings. A dimension with no opportunity in the trace is marked not applicable rather than scored, so channel repair on a clean trace earns nothing and costs nothing.

\subsection{Points}

Every applicable action and criterion has an importance $w_i\in\{1,2,3\}$, and its label maps to $r_i\in\{0,0.5,1\}$ for not met, partial, and met. Conversation ratings $g_i\in\{1,\ldots,5\}$ map to $r_i=(g_i-1)/4$. Not-applicable items enter neither side of
\begin{equation}
  S_P=100\,\frac{\sum_{i\in P}w_ir_i}{\sum_{i\in P}w_i},
  \label{eq:pillar}
\end{equation}
computed per pillar $P$ and averaged into the overall score of Equation~\ref{eq:score-overview}. Weighting the pillars equally is a design choice rather than an empirical finding. Every report exposes both pillars and their raw point totals so a reader who disagrees can reweight them.

Underneath the pillars, reports break the same points into the ten diagnostics of Table~\ref{tab:diagnostics}. Each names one observable behavior. They explain where an aggregate came from, but they are not separate objectives and they do not sum to 100.

Confidence intervals come from 2,000 hierarchical bootstrap samples that redraw agents, then scenarios within agents, then repeats. Ten scenarios that share one long operating prompt are not ten independent samples, and a flat bootstrap would pretend they are. Paired comparisons resample within-scenario score differences on the cases both models completed. The provisional success cutoff of 70 is uncalibrated, and none of our conclusions depend on it.

\FloatBarrier
\section{Checking the judges}
\label{sec:validation}

A detailed judging prompt does not by itself make a judge reliable. The action checker needs no defense (it is code running on typed events), but the two judges do. Our approach is to let code enforce everything code can check and then probe the semantic decisions no check can reach.

Code enforces the output contract described above, and it also fixes applicability. When a criterion's applicability follows from scenario data or from the deterministic action results, code resolves it and the judge must agree. Only conditions that genuinely need reading, such as whether the caller changed her mind, are left to the judge. When several judge samples are drawn per trace, the kept value is the lower median of the emitted labels, so the aggregate always lands on an anchor a judge actually produced and ties resolve conservatively. The stored minimum, maximum, and spread let a reader separate judge noise from model noise. The reference study below uses one sample per trace and cannot estimate that noise.

\begin{table}[t]
  \caption{Controlled evaluator probes. Conversation values use the 1--5 anchor scale. Each observed response is the median across three judgments.}
  \label{tab:counterfactuals}
  \centering
  \footnotesize
  \begin{tabularx}{\linewidth}{@{}P{0.25\linewidth}YP{0.31\linewidth}@{}}
    \toprule
    Planted change & Intended distinction & Observed response \\
    \midrule
    Wrong tool argument & Attempt versus correct execution & action met $\rightarrow$ partial, outcome unchanged \\
    Missing consent & Tool success versus allowed action & rule met $\rightarrow$ not met \\
    Truthful tool failure & Good recovery versus achieved state & outcome met $\rightarrow$ partial, grounding unchanged \\
    Fabricated success & Fluent claim versus supported claim & grounding met $\rightarrow$ not met \\
    Mid-call language switch & Understandable versus language-consistent & $5\rightarrow3$ \\
    Romanized target language & Right language versus TTS-ready script & $4\rightarrow2$ \\
    Repeated answered question & Local fluency versus retained state & $3\rightarrow2$ \\
    Split identifier mishandled & Fragments versus integrated value & $5\rightarrow2$ \\
    Corrupted identifier accepted & Observed text versus repaired value & $5\rightarrow3$ \\
    \bottomrule
  \end{tabularx}
\end{table}

The most direct probe of a grader is a pair of traces that differ in exactly one behavior. We built nine such pairs, each with a clean version and a defect version, and judged every trace three times. Four task-side pairs exercised the checker and the scenario judge and produced all 36 expected labels. Five conversation pairs each moved the targeted rating, and Table~\ref{tab:counterfactuals} lists them. All 15 defect ratings matched their intended anchor exactly. Two of the 15 clean-side ratings came out one point from intention, and five of the twelve untargeted conversation ratings also moved, which is partly expected since one behavior can touch two dimensions, and partly because these probes predate the final ownership rules. These tests establish that the evaluator responds to the right thing locally. They do not measure its accuracy over the full benchmark.

We also sampled twelve of the judges' unscored rule-violation flags, three per reference run, and read each against its trace. All twelve held up, among them a disclosure made before identity verification and a claim that a record was saved when no tool return supports it. Twelve samples are far too few to estimate precision, and the audit was not blind.

So the deterministic action results are verified facts, while the scenario and conversation scores remain model estimates that no blinded human rating has yet confirmed. A proper calibration study would have human raters label the same calls without knowing which model produced them, cover every language in the benchmark, and report how often the judges agree with those labels and how much repeated judgments move \citep{bavaresco2025llmjudges,fu2025multilingualjudge,haldar2025ratingroulette,purwar2026voicejudge}.

\section{Results}
\label{sec:experiments}

\subsection{Setup}

We evaluated seven candidates on the same bundle. \model{gpt-5.6-sol}, \model{electron}, \model{gpt-5.2}, and \model{phonellm} ran first. \model{gemini-3.6-flash}, \model{gpt-5.6-luna}, and \model{qwen3.8-27b} were added later, and their saved conversations were judged afterward in a regrade pass. Every run used \model{gpt-5.2-2025-12-11} without reasoning as the caller and \model{claude-opus-5} as both judges. Each case got one rollout, one judgment per judge, and at most thirty turns. Candidate reasoning was disabled wherever the endpoint allowed it, which for \model{gemini-3.6-flash} means the lowest available setting rather than off. \model{electron} ran at temperature 0.7 with top-$p$ 0.9 and top-$k$ 20, \model{phonellm} at temperature 0 with thinking disabled, and the API candidates did not support seeds. All manifests share bundle hash \texttt{04576f4b3a99...}, and eight to nineteen cases per run were dropped, most after persistent caller validation failures. The runs are marked non-official and the first four record a dirty working tree at \texttt{3fd5adc...}, so a public leaderboard will need clean tagged reruns with repeats.

Skipped episodes are excluded from means rather than scored zero, because a caller-simulator failure says nothing about the candidate. Paired comparisons use only the scenarios both models completed and resample the within-scenario differences. The reported intervals cover benchmark sampling only. Caller randomness, judge randomness, and the authors' importance weights all sit outside them.

\subsection{Scores}

\begin{table*}[t]
  \caption{Reference results. Actions and Criteria are the deterministic and judged halves of Task. Intervals are 95\% hierarchical bootstrap intervals over agents and scenarios, and Pass \% uses the provisional cutoff of 70. Each scored scenario has one rollout and one judgment per judge.}
  \label{tab:main-results}
  \centering
  \footnotesize
  \begin{tabular}{@{}lrrrrrrrr@{}}
    \toprule
    Candidate & Scored & Actions & Criteria & Task & Conversation & Overall & 95\% interval & Pass \% \\
    \midrule
    \model{gpt-5.6-sol} & 481 & 63.2 & 82.7 & 73.6 & 74.1 & \textbf{73.9} & [71.2, 76.3] & 65 \\
    \model{electron} & 480 & 70.8 & 72.5 & 71.6 & 68.1 & 69.8 & [67.2, 72.4] & 57 \\
    \model{gemini-3.6-flash} & 471 & 68.3 & 64.8 & 67.9 & 56.6 & 62.3 & [59.7, 64.9] & 34 \\
    \model{gpt-5.6-luna} & 473 & 58.6 & 66.2 & 62.4 & 54.2 & 58.3 & [55.8, 60.8] & 22 \\
    \model{gpt-5.2} & 482 & 65.5 & 59.0 & 62.9 & 44.9 & 53.9 & [51.8, 56.0] & 10 \\
    \model{qwen3.8-27b} & 475 & 64.2 & 62.2 & 62.8 & 36.3 & 49.5 & [46.5, 52.5] & 12 \\
    \model{phonellm} & 482 & 28.2 & 41.2 & 35.2 & 31.6 & 33.4 & [30.0, 37.1] & 3 \\
    \bottomrule
  \end{tabular}
\end{table*}

\begin{figure*}[t]
  \centering
  \includegraphics[width=\textwidth]{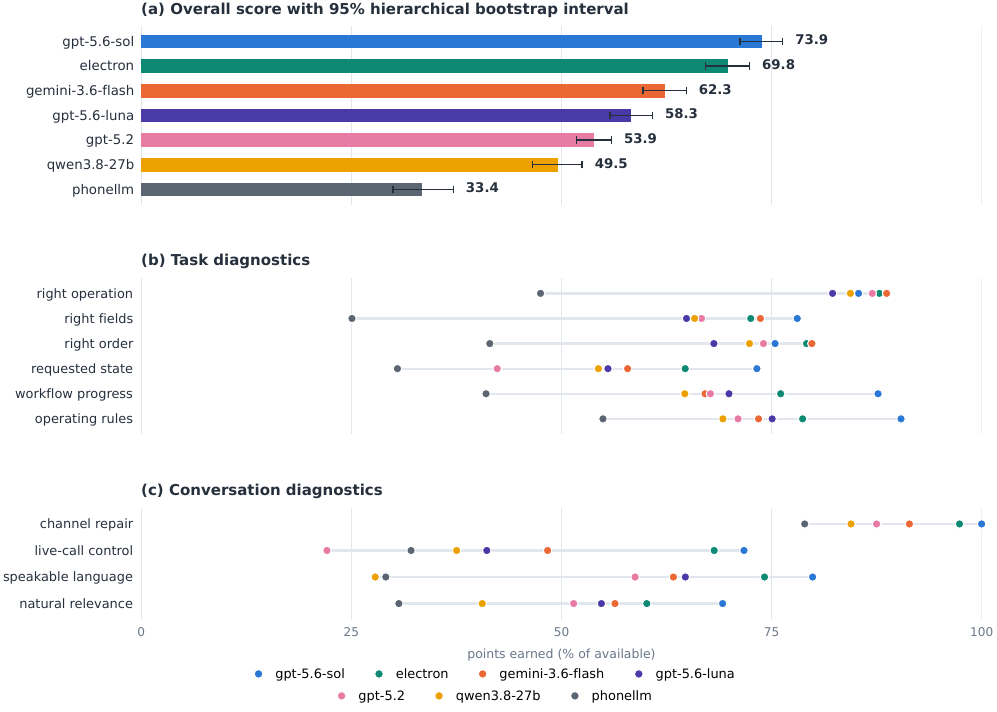}
  \caption{Reference results from saved artifacts. Panel (a) shows the overall score with its 95\% interval. Panels (b) and (c) show the share of importance-weighted points earned on each diagnostic, one dot per candidate, with a gray line spanning the range across candidates. All three panels share the horizontal scale.}
  \label{fig:results}
\end{figure*}

\begin{table*}[t]
  \caption{Mean overall score by authored language and difficulty. Counts vary slightly because caller-invalid episodes are excluded. Language slices contain different agents and tasks, so they are not controlled estimates of a language effect.}
  \label{tab:slices}
  \centering
  \footnotesize
  \setlength{\tabcolsep}{4.2pt}
  \begin{tabular}{@{}lrrrrrrrrr@{}}
    \toprule
    Candidate & English & Gujarati & Hindi & Marathi & Spanish & Tamil & Telugu & Medium & Hard \\
    \midrule
    \model{gpt-5.6-sol} & 79.5 & 76.9 & 67.0 & 74.7 & 70.9 & 70.6 & 75.5 & 75.3 & 72.5 \\
    \model{electron} & 77.6 & 70.0 & 67.7 & 71.5 & 65.7 & 66.9 & 67.8 & 71.8 & 67.8 \\
    \model{gemini-3.6-flash} & 69.1 & 64.7 & 58.7 & 62.1 & 58.8 & 58.6 & 62.3 & 64.3 & 60.2 \\
    \model{gpt-5.6-luna} & 62.8 & 57.9 & 52.9 & 60.7 & 54.1 & 57.5 & 60.5 & 59.6 & 56.9 \\
    \model{gpt-5.2} & 57.2 & 54.0 & 47.9 & 58.8 & 52.5 & 54.6 & 50.9 & 56.0 & 51.8 \\
    \model{qwen3.8-27b} & 62.5 & 45.6 & 45.9 & 48.5 & 52.0 & 40.7 & 48.8 & 51.8 & 47.3 \\
    \model{phonellm} & 52.9 & 26.1 & 33.4 & 22.4 & 37.4 & 30.4 & 28.6 & 34.2 & 32.6 \\
    \bottomrule
  \end{tabular}
\end{table*}

\begin{table}[t]
  \caption{Adjacent models compared on the scenarios both completed. Differences are mean within-scenario gaps in overall score, with 95\% bootstrap intervals.}
  \label{tab:pairwise}
  \centering
  \footnotesize
  \begin{tabular}{@{}lrr@{}}
    \toprule
    Pair & Difference & 95\% interval \\
    \midrule
    \model{gpt-5.6-sol} vs \model{electron} & +4.1 & [1.7, 6.5] \\
    \model{electron} vs \model{gemini-3.6-flash} & +7.8 & [5.9, 9.7] \\
    \model{gemini-3.6-flash} vs \model{gpt-5.6-luna} & +4.0 & [2.1, 5.8] \\
    \model{gpt-5.6-luna} vs \model{gpt-5.2} & +4.3 & [2.2, 6.4] \\
    \model{gpt-5.2} vs \model{qwen3.8-27b} & +4.5 & [1.8, 7.2] \\
    \model{qwen3.8-27b} vs \model{phonellm} & +16.1 & [13.2, 19.0] \\
    \bottomrule
  \end{tabular}
\end{table}

Table~\ref{tab:main-results} and Figure~\ref{fig:results} summarize the runs. \model{gpt-5.6-sol} leads at 73.9 overall, and even it clears the provisional cutoff in only 65 percent of its cases. Because every candidate ran on the same scenarios, adjacent models can also be compared pairwise on the cases both completed, which removes most of the scenario-to-scenario variance. Table~\ref{tab:pairwise} lists these comparisons. Every difference is positive and every interval excludes zero, so the ordering in Table~\ref{tab:main-results} holds even where its marginal intervals overlap. The closest pair is at the top, where \model{gpt-5.6-sol} wins 295 of the 475 cases it shares with \model{electron}, loses 157, and ties 23.

\subsection{What the results suggest}

\paragraph{Picking the right tool is not what separates models.}
Six of the seven models choose the correct operation at nearly the same rate, between 82.3 and 88.7 percent of available points, yet their overall scores span 24.4 points. The model with the best operation choices, \model{gemini-3.6-flash}, finishes third. The separation comes from what surrounds the call. Across those six models, correct argument values range from 64.9 to 78.1, and reaching the state the caller actually asked for ranges from 42.4 to 73.3. A benchmark that stopped at tool names would call these six models roughly equal.

\paragraph{Task and conversation fail independently.}
The three models in the middle of the table are within half a point of one another on task score and spread over 18 points on conversation. Their traces point to two recurring behaviors. The first is silence around tool calls, counting a call as silent when its message carries no spoken text and no concluding message. \model{gpt-5.2} is silent on 2,672 of its 2,765 tool calls and ends 460 of its 461 model-ended calls without a word, where \model{electron} is silent on roughly one call in ten. From the caller's side the line simply goes quiet. The second is script. \model{qwen3.8-27b} wrote mostly Latin script in 43 of its 60 Hindi conversations, and a Hindi TTS voice cannot speak romanized text as intended. Neither behavior touches the task score, but a caller would notice both immediately.

\paragraph{Some failures never reach the score.}
The scenario judge also flags rule violations that no authored criterion covers, and these flags stay out of the score (Section~\ref{sec:evaluation}). Their volume is still worth reporting. Across the seven runs the judge raised between 162 and 1,051 flags per model, and even the second-place model drew flags in more than half of its calls. These are capped, model-judged counts rather than verified error rates, but they indicate that the authored criteria do not cover everything that goes wrong, and that some consent and disclosure failures never appear in the headline number.

\paragraph{Channel evidence is thin.}
Only 32 to 40 scored traces per run activated channel handling. Individually those cases are informative, but they cannot support general claims about how a model handles recognition errors, which would need more authored damage across languages and value types than this bundle carries.

\subsection{Language and difficulty}

Table~\ref{tab:slices} splits the same scores by language and difficulty. Every model scores lower on hard cases than on medium ones, and every model except \model{gpt-5.2} scores highest in English. The two weakest models also have the widest language spreads. \model{qwen3.8-27b} drops more than 20 points from English to Tamil, in line with its script problem, and \model{phonellm} drops 30 from English to Marathi. We resist reading these slices as language effects, because each slice contains different agents and tasks. A causal comparison would need the same scenarios translated and independently checked for equivalence.

We read the overall score as a profile rather than a ranking. A model change can improve task completion while making tool transitions harder to follow, or polish the speech while weakening rule compliance. Reporting two pillars and ten diagnostics keeps trades of that kind visible instead of averaging them away.

\section{Related work}

\paragraph{Task dialogue and tool use.}
MultiWOZ and the Schema-Guided Dialogue dataset established multi-domain state tracking and response evaluation \citep{budzianowski2018multiwoz,rastogi2020sgd}. $\tau$-bench, $\tau^2$-bench, and ToolSandbox evaluate stateful interaction among users, policies, tools, and environments \citep{yao2024taubench,barres2025tau2,lu2025toolsandbox}. CONFETTI studies turn-level function use, MASSIVE-Agents extends function calling across languages, and MultiChallenge and recent customer-service benchmarks stress long instructions and extended dialogue \citep{alkhouli2025confetti,kulkarni2025massiveagents,deshpande2025multichallenge,gao2026olabench}. \bench adds the obligations a voice channel creates in text and separates exact execution from spoken conduct.

\paragraph{Speech and voice agents.}
VoiceBench evaluates spoken knowledge, instruction following, and safety, and Audio MultiChallenge and MTalk-Bench study natural multi-turn spoken interaction \citep{chen2026voicebench,gosai2026audiomultichallenge,du2025mtalkbench}. VoiceAgentBench adds multilingual tool workflows \citep{jain2025voiceagentbench}. EVA-Bench, $\tau$-Voice, and VAmoS Bench evaluate complete voice systems with audio, task state, or duplex behavior \citep{bogavelli2026evabench,ray2026tauvoice,meyer2026vamos}, and Full-Duplex-Bench-v2 and Moshi study simultaneous spoken dialogue directly \citep{lin2026fullduplex,defossez2024moshi}. Those benchmarks measure the layers \bench deliberately leaves out. An end-to-end score describes the whole product, and \bench attributes behavior to the language model alone.

\paragraph{Model-based evaluation.}
G-Eval showed that explicit criteria and structured reasoning improve model-based text evaluation \citep{liu2023geval}. HealthBench grades conversations against physician-written criteria and compares model grading with physician grading \citep{arora2025healthbench}. Later studies find that judge quality varies with task, language, prompt, and repetition \citep{bavaresco2025llmjudges,fu2025multilingualjudge,haldar2025ratingroulette}. PaperBench separates executable checks from rubric judgment and validates its judge against expert labels \citep{starace2025paperbench}. \bench limits judge authority in a similar spirit, but our probes and audits are not a substitute for an expert-labeled judge study.

\section{Limitations}
\label{sec:limitations}

\paragraph{No audio.}
The benchmark measures nothing about sound. Recognition error rates, TTS quality, pronunciation, accent, emotion in the voice, background noise, overlapping speech, and latency are all out of scope, and text-level damage stands in for only a narrow slice of what real pipelines do. The results cannot rank complete voice systems.

\paragraph{A narrow model of turn taking.}
VAD splits arrive as consecutive messages before one model turn. They test whether fragments are read as one meaning, not endpoint timing, interruptions, or barge-in, which need duplex evaluation.

\paragraph{Simulated callers and backends.}
An LLM caller and ordered fixtures give control at the cost of realism. Real callers hesitate on their own schedule, change goals, and hit concurrent backends with partial failures. One caller model may also favor candidates that resemble it. External validity needs human-call and live-backend studies.

\paragraph{Judge calibration.}
Typed evidence rules out invented citations, not wrong readings. The reference runs used one judgment per trace, there is no blinded human comparison yet, and judge bias by candidate family is unmeasured. Small gaps between models, and the language slices, should be read loosely.

\paragraph{Corpus coverage.}
The 49 agents were authored, not sampled from deployment traffic, and the languages are not balanced for dialect, region, or code mixing. Only 53 scenarios author channel damage. Public cases can leak into future training data, so held-out bundles and contamination checks will be needed.

\paragraph{Weights and variance.}
Equal pillar weights and the 1-to-3 importance weights were set by the authors and may not match any given deployment's cost of failure. The study has one rollout per scenario, drops eight to nineteen cases per model, and ran from development revisions rather than a tagged release. It establishes differences within this recorded study, not stable provider-wide rankings.

The scenarios include medical, financial, and emergency settings because those are the calls where verification, disclosure, and grounded claims matter most. A benchmark score neither authorizes deployment in such a setting nor replaces an application's own safety review.

\section{Conclusion}

\bench measures the language model that sits between a voice pipeline's recognizer and its speech output. It simulates the caller, the channel, and the backend around the model, and it grades the saved call with three evaluators that each see only the evidence their decision needs. In our reference study, tool selection, the measure most agent benchmarks report, barely separated the stronger models. The differences came from argument values, ordering, rule compliance, the state the caller actually reached, and what the model said around its tool calls. Three pieces of work would make these scores more trustworthy. The judges should be checked against human raters who do not know which model produced each call, in every language the benchmark covers. The reference runs should be repeated from a tagged release. And the corpus needs many more scenarios with authored recognition damage, since the current 53 are too few to support claims about it.

\balance
{\small
\setlength{\bibsep}{3pt plus 0.5pt}
\bibliographystyle{plainnat}
\bibliography{references}
}

\end{document}